# On the Parametric Study of Lubricating Oil Production using an Artificial Neural Network (ANN) Approach

Masood Tehrani[1] and Mary Ahmadi[2]
[1]Department of Engineering, NJIT, Newark, NJ, USA 07102
[2]Depratment of Information Systems, Think box, Irvine, CA, USA

**Abstract**

In this study, an Artificial Neural Network (ANN) approach is utilized to perform a parametric study on the process of extraction of lubricants from heavy petroleum cuts. To train the model, we used field data collected from an industrial plant. Operational conditions of feed and solvent flow rate, Temperature of streams and mixing rate were considered as the input to the model, whereas the flow rate of the main product was considered as the output of the ANN model. A feed-forward Multi-Layer Perceptron Neural Network was successfully applied to capture the relationship between inputs and output parameters.

## 1. Introduction

It is necessary to remove aromatic compounds from heavy petroleum cuts to improve the quality of the produced lubricating base oils [1-2]. Extraction of aromatic compounds from lubricating oil cut is usually done through a liquid-liquid (solvent) extraction process. Several works have been done to find a suitable solvent for extraction of aromatic hydrocarbons from lube-oil cut [3-5]. Sulfolane is heavily used as solvent for this process due to high selectivity for aromatics extraction. This compound is suitable to extract light and heavy vacuum distillates [6-9]. In a liquid-liquid (solvent) extraction process, two streams of feed and solvent need to be kept in contact with one another within the column. Studies on the efficiency of conventional extraction contactors show that RDC (Rotating Disc Contactors) columns can be considered as the most efficient extraction column [10-13]. Research has been focused on predicting the efficiency of RDC column with respect to the physical properties of the inlet streams as well as the geometry and other characteristics of the system [14]. Performance of RDC columns has not been accurately predictable due the complexity of heat and mass transfer taken place in these components. Acritical Neural Network (ANN) can be used to model processes in different areas of chemical engineering [15-22] especially in the function approximation. Function approximation is based upon the training of an ANN against sets of input–output data pairs in an attempt to determine the relationship between the input and output parameters. In this study, an ANN model is used to develop simulation framework for the parametric study of an RDC column with which the solvent extraction process for production of lubricating oils takes place.

## 2. Materials and Methods

### 2.1 *Chemical Extraction Process*

One of the ways to produce lubricating oils is through solvent extraction processes in which the feed, normally a lubricating oil cut, will get in contact with an extraction solvent, here Sulfolane [23]. In this paper industrial field data were used to train an ANN model capable of predicting the performance of the column.

### 2.2 *Neural Network Architecture*

A Feed-Forward Multilayer Perceptron Neural Network was built and got trained using a large sets of data as the input and output parameters of the model. These parameters were carefully selected to represent the real system. The input parameters were independent from one and other and were the ones which most affect the flow rate of the main product, including the temperature of feed and solvent, rate of rotation of discs and solvent to feed ration and the only output parameter was selected to be the flow rate of main product stream. Characteristics of the ANN model are tabulated in Table1.

Table 1. Inputs and Output parameter of the ANN model

| Inputs | Outputs |
| --- | --- |
| I= solvent/feed ratio | Product flow rate ($m^3$/hr) |
| Feed temperature ($^0$C) | |
| Solvent temperature ($^0$C) | |
| Rotation rate (rpm) | |

Cybenko [17] shows that for an ANN model, one hidden layer can be adequate as long as a right number of nodes are selected to predict the transition function. Fig.1 depicts the architecture of the ANN model used in this study.

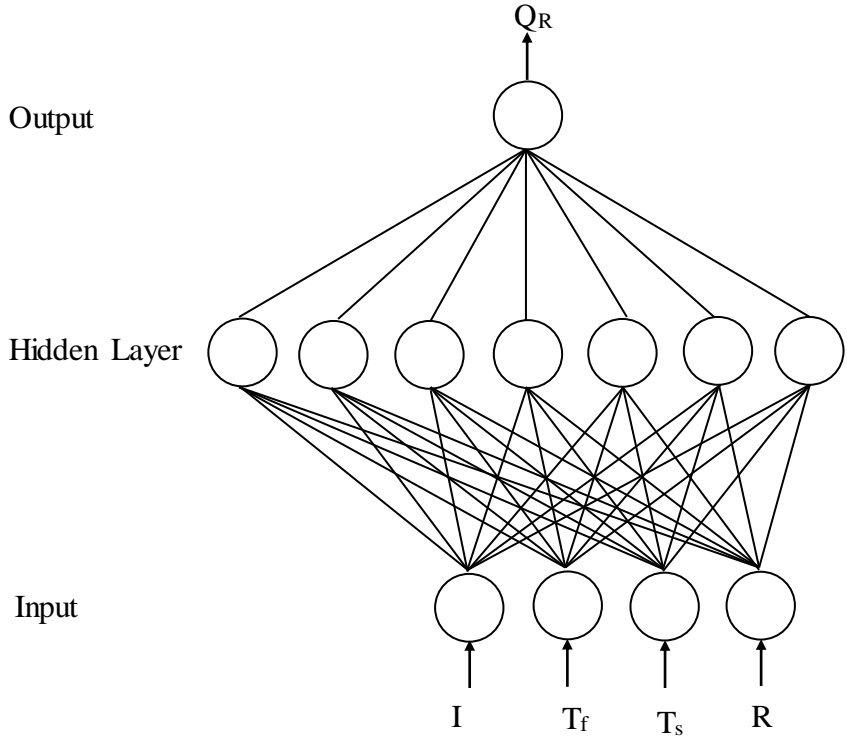

Fig 1. Schematic of the ANN model

In this study, 400 data sets were used to develop the model from which 80% were participated in the training process, and the rest were used to validate the trained model.

A Multi-Layer Perceptron Neural Network (MLPNN) model with 7 nodes in the hidden layer showed best results. A sigmoid activation function was used for the hidden layer with the following form:

$$f(x) = \frac{1}{1+e^{(-x-bias)}}$$

The back-propagation error was used as an index to train the network. Two values were used to measure the performance of the ANN to find the best model for a particular data set as the following:

1) % Error

Defined by the following formula:

$$\%Error = \frac{100}{N'P}\sum_{j=0}^{P}\sum_{i=0}^{N'}\frac{|dy_{ij}-dd_{ij}|}{dd_{ij}}$$

where P = number of output elements

N' = number of exemplars in the data set

$dy_{ij}$ = denormalized output for exemplar (i) at processing element (j)

$dd_{ij}$ = denormalized desired output for exemplar (i) at processing element (j)

2) MSE: (Mean Square Error)

$$MSE = \frac{\sum_{j=0}^{P} \sum_{i=0}^{N'}(d_{ij} - y_{ij})^2}{N'P}$$

Where P = number of processing elements

N' = number of exemplars in data set

$y_{ij}$ = network output for exemplar (i) at processing element (j)

$d_{ij}$ = desired output for exemplar (i) at processing element (j)

## 3. Results and Discussion

### 3.1 ANN modeling/Pattern Generation

The optimum number of neuron nodes in the hidden layer, was selected based on the minimum values of the (MSE) and (Error%) of different ANN models created using different number of nodes. Table 2 lists these parameters after 100000 iterations of the model. Looking at the results, it was found out that a network with 7 hidden nodes has the lowest values of MSE and Error% so best suits our purposes.

Table 2. Characteristics of the optimized ANN

| Number of nodes in hidden layer | MSE | % Error |
|---|---|---|
| 7 | 0.034 | 2.854 |

The average relative error and the maximum relative error for the outputs of the network after training the 4-7-1 network are listed in Table 3:

Table 3. Values of the errors in the trained ANN model

| | Average Relative Error (%) | Maximum Relative Error (%) |
|---|---|---|
| Flow rate of product (m³/hr) | 0.65 | 1.05 |

To further validate the accuracy of the ANN model, 80 data points which were not participated in the modeling process were used for comparison purposes. As shown in Fig.2 the predicted data points using ANN model are very close to the actual values.

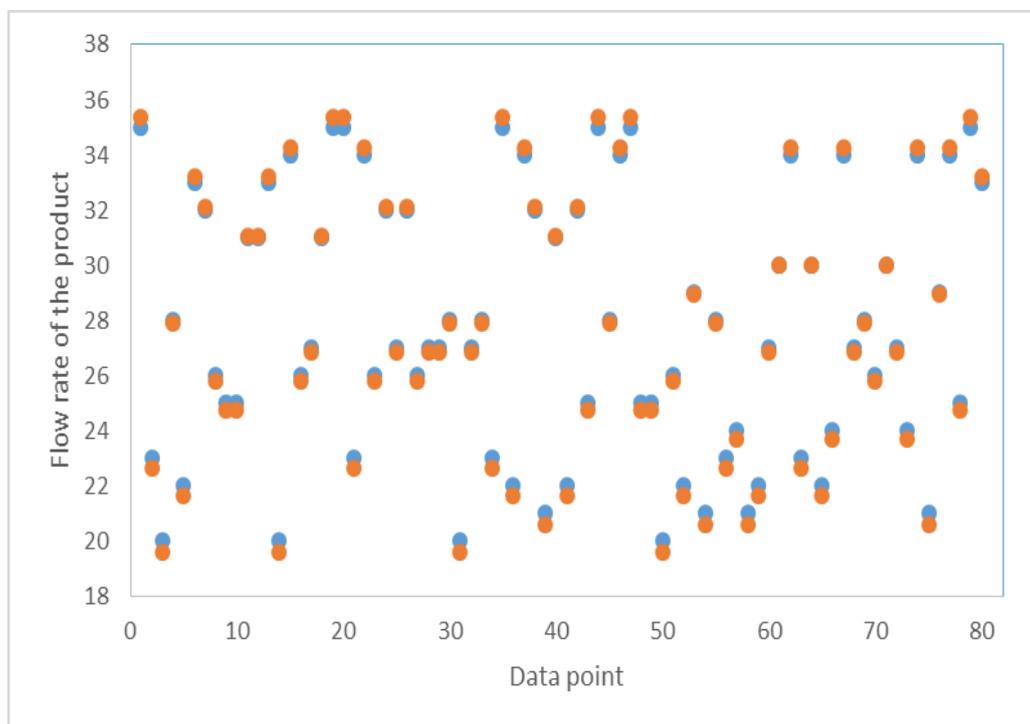

Fig 2. Actual vs. model predicted flow rate of the main product

### 4. Conclusion

- It was possible to safely perform a parametric study on the performance of a RDC column with respect to its operating conditions using an Artificial Neural Network procedure.
- The results have shown that by increasing the solvent to feed ratio, the amount of aromatic compounds extracted from the feed stream will be increased.
- Increasing the rotation rate will enhance the extraction process.